\documentclass[runningheads]{llncs}

\usepackage[preprint]{eccv}

\usepackage{eccvabbrv}
\usepackage{euscript}
\usepackage{multirow}
\usepackage{multicol}
\usepackage[table]{xcolor}
\usepackage{pifont}
\usepackage{threeparttable}
\usepackage{booktabs}
\usepackage{mathrsfs}

\usepackage{threeparttable}   % 提供 threeparttable 环境和 tablenotes
\usepackage{graphicx}         % 提供 \resizebox 命令
\usepackage{makecell}         % 提供 \makecell 命令（用于单元格内换行）
\usepackage{booktabs}         % 提供 \toprule, \midrule, \bottomrule
\usepackage{algorithm}
\usepackage{algpseudocode}
\usepackage{amsmath} 

\usepackage[accsupp]{axessibility}  % Improves PDF readability for those with disabilities.

\usepackage[pagebackref,breaklinks,colorlinks,citecolor=eccvblue]{hyperref}
\usepackage{xcolor}
\usepackage{orcidlink}

\begin{document}

% \title{\textit{What a Nice Agent!} 
\title{\textit{ \textcolor{green!60!black}{Evolving}} Medical Imaging Agents via
\textit{\textcolor{orange!55!white}{Experience}}-driven Self-skill Discovery}
%MedAgentSkill: Learning to Discover and Compose Visual Skills for Medical Image Understanding

%MedAgentSkill: Self-Skill Discovery for Medical Vision-Language Agents

\titlerunning{Evolving Medical Imaging Agents via Experience-driven Self-skill Discovery}

% Include the authors' ORCID for the camera-ready version, if at all possible.
\author{Lin Fan\thanks{Lin Fan and Pengyu Dai are contributed equally.}\inst{1}\orcidlink{0000-0003-1198-9582} \and
Pengyu Dai\protect\footnotemark[1]\inst{2,3}\orcidlink{0009-0006-3705-4418} \and
Zhipeng Deng\inst{4}\orcidlink{0009-0005-7602-5460} \and
Haolin Wang\inst{5}\orcidlink{0000-0002-5767-0019} \and
Xun Gong\inst{1}\orcidlink{0000-0002-1494-0955} \and
Yefeng Zheng\inst{4}\orcidlink{0000-0003-2195-2847} \and
Yafei Ou\thanks{Corresponding Author (\email{yafei.ou@riken.jp})}\inst{3}\orcidlink{0000-0001-7510-0813}}

\authorrunning{L.~Fan et al.}

\institute{Southwest Jiaotong University, China \and
The University of Tokyo, Japan \and
RIKEN, Japan\and
Westlake University, China\and
Hokkaido University, Japan}

\maketitle

\begin{abstract}
Clinical image interpretation is inherently multi-step and tool-centric: clinicians iteratively combine visual evidence with patient context, quantify findings, and refine their decisions through a sequence of specialized procedures. While LLM-based agents promise to orchestrate such heterogeneous medical tools, existing systems treat tool sets and invocation strategies as static after deployment. This design is brittle under real-world domain shifts, across tasks and evolving diagnostic requirements, where predefined tool chains frequently degrade and demand costly manual re-design.
We propose MACRO, a self-evolving, experience-augmented medical agent that shifts from static tool composition to experience-driven tool discovery. From verified execution trajectories, the agent autonomously identifies recurring effective multi-step tool sequences, synthesizes them into reusable composite tools, and registers these as new high-level primitives that continuously expand its behavioral repertoire. A lightweight image-feature memory grounds tool selection in a visual–clinical context, while a GRPO-like training loop reinforces reliable invocation of discovered composites, enabling closed-loop self-improvement with minimal supervision.
Extensive experiments across diverse medical imaging datasets and tasks demonstrate that autonomous composite tool discovery consistently improves multi-step orchestration accuracy and cross-domain generalization over strong baselines and recent state-of-the-art agentic methods, bridging the gap between brittle static tool use and adaptive, context-aware clinical AI assistance. Code will be available upon acceptance.
% AI agents integrating Large language models (LLMs) with specialized tools offer a path toward balancing generality and precision. However existing systems rely on static tool sets and fixed invocation strategies, leaving them vulnerable to performance degradation under domain shifts.
% To address this limitation, we introduce a self-evolving, experience-augmented agent framework that autonomously discovers novel composite tool sequences from verified executions, dynamically expanding its library of reusable high-level primitives through accumulated experience. Rather than relying on manually predefined workflows, the system continuously learns which multi-step tool chains are practically relevant and adapts its behavior to invoke them adaptively under real-world variability. By integrating a lightweight image-feature memory that grounds tool selection in clinical context, and a reward-aligned training loop that reinforces reliable invocation of registered composites, the agent achieves closed-loop self-improvement with minimal manual supervision.
% We validate the framework across diverse medical imaging datasets, benchmarking against both established baselines and state-of-the-art methods, demonstrating that autonomous composite discovery bridges the gap between static tool use and adaptive, context-aware assistance in clinical AI.  
\keywords{Medical AI Agents
\and Agent Self-Evolving
\and Tool Integrated Reasoning 
}
\end{abstract}

\section{Introduction}
\label{sec:intro}

\begin{figure}[!ht]
  \centering
  \includegraphics[width=\textwidth]{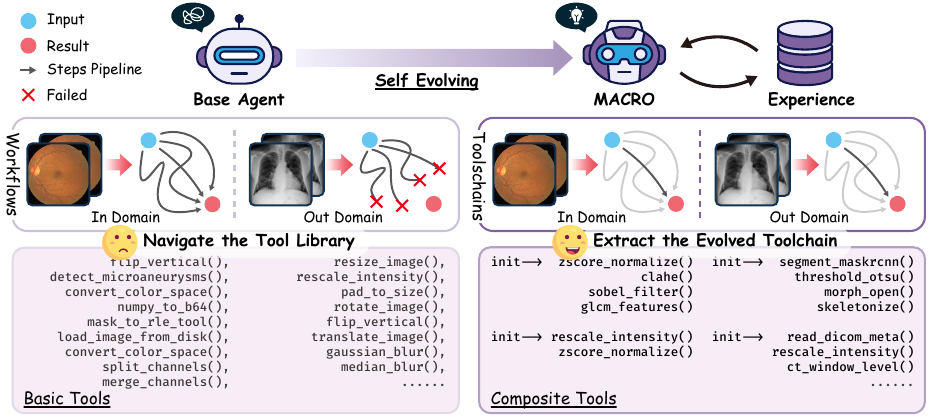}
  \caption{Existing medical agents vs. our MACRO. Existing medical agents rely on static, predefined tool sets and invocation strategies that struggle to adapt to variations across different imaging domains or intra-class sample differences, often leading to failures. In contrast, MACRO   enhances adaptability by discovering and integrating composite tools, validated tool sequences capable of executing multi-step operations. These sequences are distilled from repeatedly successful trajectories in real-world workflows, enabling more robust performance in the face of clinical variability.}
  \label{fig:motivation}
\end{figure}
% LLM问题

Medical image interpretation is rarely a single-step prediction. In real clinical workflows, radiologists iteratively combine visual evidence with patient context, compare against priors, consult guidelines, and invoke a sequence of specialized procedures (e.g., detection, segmentation, quantification, and longitudinal tracking) before making a decision. Such workflows are inherently \emph{multi-step}, \emph{tool-centric}, and \emph{verification-driven}. In contrast, most existing evaluations of large language models (LLMs) in medicine focus on isolated tasks or single-turn question answering, which under-represents the planning, interaction, and contextual variability that dominate real-world practice~\cite{zhao2023survey,moor2023foundation,tu2024towards}. This mismatch raises a central concern: even if general-purpose LLMs appear strong on curated benchmarks, their reliability and rigor under realistic clinical complexity remain unclear~\cite{derraz2024new}.
% 进化为使用工具

Meanwhile, measurable progress in precision medicine has largely been driven by \emph{specialized} deep learning systems engineered for particular modalities or subtasks~\cite{kather2019deep}, offering strong accuracy but limited flexibility and high maintenance cost when workflows shift across centers, scanners, or requirements. A natural direction is therefore to combine the \emph{planning and language interface} of tool-enabled LLM agents~\cite{yao2022react,schick2023toolformer,tayebi2024large} with the \emph{domain fidelity} of medical vision/analytics models~\cite{messiou2023multimodal}. Recent agentic systems indeed show that coupling LLMs with external tools and APIs can improve complex task solving~\cite{liu2023agentbench,liang2024encouraging,chen2023communicative,wang2024beyond}. Meanwhile, medical variants explore multi-agent clinical reasoning or integrate imaging tools such as segmentation and detection modules~\cite{kim2024mdagents,tang2024medagents,liu2024medchain,fallahpour2025medrax,li2024mmedagent}.

% 现有使用工具的LLM存在的问题
However, existing medical agents share a limiting design assumption: \emph{tool sets and invocation strategies are predefined and remain static after deployment}. This assumption is brittle in clinical reality. Hospitals differ in acquisition protocols, reporting templates, and data formats; patient distributions drift; imaging modalities and diagnostic requirements evolve. Under such domain shifts, agents built on fixed tool chains and scripted orchestration often degrade sharply and require extensive manual re-engineering to restore interoperability and performance~\cite{mehandru2024evaluating,tu2025towards,liu2024medchain,almansoori2025medagentsim,ajitha2025medxagent}. In other words, today’s systems can be competent \emph{tool users}, yet remain poor \emph{tool learners} (as shown in Fig.~\ref{fig:motivation}): they lack a mechanism to autonomously discover, validate, and internalize new multi-step routines that repeatedly prove useful in practice.

We argue that robustness in medical imaging assistance demands a paradigm shift: \textit{from static tool composition to experience-driven tool discovery}. Clinicians do not operate by re-solving each case from scratch; instead, they accumulate experience into reusable diagnostic routines (e.g., a consistent sequence for lesion measurement, cross-slice verification, and longitudinal comparison), refining these routines as new cases and constraints arise. Translating this principle to agentic systems suggests that an effective medical agent should be able to \emph{grow its own repertoire} of reliable multi-step primitives from interaction experience, and reuse them as high-level actions when encountering new cases.

To this end, we introduce a self-evolving, experience-augmented medical agent: \textbf{M}edical \textbf{A}gent for \textbf{C}omposite \textbf{R}easoning and \textbf{O}rchestration (MACRO), that continuously expands its capabilities by discovering and incorporating \emph{composite tools}---verified multi-step tool sequences distilled from recurring successful trajectories in real workflows. Concretely, our framework couples (i) an \emph{image-feature memory} that grounds tool selection in visual--clinical context, with (ii) a \emph{composite tool synthesis} module that autonomously identifies, validates, and registers reusable multi-step sequences as new high-level primitives. A reward-aligned training loop then reinforces reliable invocation of these synthesized composites, enabling closed-loop self-improvement with minimal manual supervision. By turning experience into reusable procedural skills, the agent becomes more adaptive under real-world variability and more consistent in multi-step clinical image interpretation.
Our contributions are threefold:
\begin{itemize}
    \item We identify a fundamental limitation of current medical AI agents: reliance on static, manually predefined tool workflows yields brittle generalization under domain shifts and imposes strong dependence on expert-designed decompositions, constraining scalability to heterogeneous and evolving clinical imaging scenarios.
    \item Inspired by clinician routine formation, we propose a self-evolving medical agent paradigm with two synergistic mechanisms: an image-feature memory for context-grounded tool selection, and a composite tool synthesis module that autonomously discovers, verifies, registers, and reuses multi-step tool sequences as high-level primitives, continuously expanding the agent’s behavioral repertoire.
    \item We validate the framework across diverse medical imaging tasks and datasets, benchmarking against established baselines and recent state-of-the-art methods. Results show that autonomous composite tool discovery consistently improves multi-step orchestration accuracy and cross-task generalization, demonstrating the effectiveness of experience-driven self-improvement for context-grounded clinical decision support.
\end{itemize}

\section{Related Works}
\subsubsection{LLM-based Medical Agent.}
Recent multimodal LLMs combine vision and language to tackle medical imaging and decision tasks~\cite{nam2025multimodal}. For example, an agent might encode a chest X-ray alongside clinical notes and generate a preliminary diagnosis or report~\cite{nam2025multimodal}. Such systems can automatically draft radiology findings or answer visual questions, promising richer clinical decision support.
Beyond single-agent models, multi-agent frameworks have been proposed for end-to-end clinical workflows. Elboardy et al~\cite{elboardy2025medical}. introduced a ten-agent system where specialized LLMs handle stages of radiology interpretation (segmentation, diagnosis, and report writing). Other architectures partition care into sequential tasks (referral → history → exam → diagnosis → treatment) with dedicated agents for each stage~\cite{liu2024medchain}. Some agents focus on subproblems (e.g., CoD ~\cite{chen2025cod} for interpretable diagnostics, EHRagent~\cite{shi2024ehragent} for medical record analysis), while others (such as Almanac Copilot~\cite{zakka2024almanac}) assist clinicians with Electronic Health Record (EHR) tasks. In all these systems, the pipeline structure and available tools are defined in advance by designers.

As a result, existing clinical agents struggle with novel or out-of-distribution cases. They typically follow rigid, step-by-step workflows and rely on static knowledge sources. For example, Liu et al~\cite{liu2026benchmarking} reported that even with powerful LLMs, agentic systems achieve only 15–30\% accuracy on unseen multimodal clinical tasks. Similarly, current medical retrieval-augmented systems use fixed chunked knowledge bases that break down on rare conditions~\cite{liu2024medchain}. In practice, a new imaging modality or unusual pathology can derail these agents. These limitations motivate a new approach: a self-evolving, experience-augmented clinical agent that autonomously discovers and integrates novel multi-step tool workflows from recurring clinical patterns\cite{huang2025biomni,wu2025evolver}.

\subsubsection{Tool-integrated Reasoning.}
Recent work has shown that large language models (LLMs) can be augmented with external tools or APIs to improve their problem-solving capabilities~\cite{wei2025beyond}. Key architectural patterns emerge in these tool-integrated LLM agents: (i) Planner–Executor separation: Some recent systems decouple planning from execution. A planner LLM generates a global plan for a complex query, and an executor then carries out each step~\cite{wei2025beyond}. This contrasts with purely reactive agents and allows parallel tool use. (ii) Iterative agent loops: Other agents use a “Thought–Action–Observation” loop. The LLM alternates between reasoning (chain-of-thought) and acting (tool calls), using observations to update its plan~\cite{yao2022react}. 

Despite their successes, existing systems face significant limitations rooted in static design. Most rely on fixed, manually curated tool libraries: Toolformer uses predetermined APIs~\cite{schick2023toolformer}, ReAct-style agents hard-code knowledge sources, and HuggingGPT assumes a static catalog of expert models~\cite{shen2023hugginggpt}. Consequently, they lack the ability to add new tools or compose novel multi-step workflows beyond what was scripted, as complex planners~\cite{wang2024beyond} rarely translate to practice. Learning from experience is also absent: agents do not update tool selection strategies based on past outcomes, missing the self-reflection emphasized for truly agentic systems~\cite{zhi2025reinventing}. Medical agents exhibit the same patterns, using fixed retrieval-augmented frameworks~\cite{garza2025retrieval} or pre-collected expert models~\cite{li2024mmedagent} that remain unchanged at run-time. This absence of adaptation and discovery from ongoing use motivates our approach: a self-evolving agent that autonomously learns and refines composite tool sequences through clinical experience.

\subsubsection{Agent Self-evolving.} 
Recent AI research has begun equipping LLM agents with mechanisms to learn from experience. Some works augment agents with memory or skill libraries that evolve over time. For instance, MemSkill~\cite{zhang2026memskill} frames memory construction as a set of reusable “skills” and alternates between applying existing skills and using an LLM to refine or create new skills from hard cases. Similarly, EvolveR distills an agent’s past interaction trajectories into a repository of abstract strategic principles and then updates the agent (via reinforcement learning) to use those principles on new tasks\cite{wu2025evolver}. These methods create closed loops where the agent gradually improves its planning and reasoning over successive experiences.
Even with these advances, most tool-augmented agents remain static in their capabilities. Off-the-shelf frameworks (e.g., agent orchestration frameworks~\cite{schick2023toolformer} and autonomous planning agents~\cite{yao2022react}) let LLMs call external APIs or tools, but only from a fixed, human-defined toolkit. Some recent protocols (such as the proposed Model Context Protocol~\cite{nam2025multimodal}) aim to let agents autonomously discover and orchestrate external tools. In biomedical research, the Biomni~\cite{huang2025biomni} agent goes further: it uses an LLM to scan thousands of research papers and build a unified action space of hundreds of tools and databases, enabling dynamic composition of complex workflows. However, even the Biomni noted that “equipping Biomni to autonomously discover and incorporate new tools and databases” is needed for long-term adaptability. 

To our knowledge, no prior healthcare AI system continuously mines its own clinical experiences to create new composite tool chains. Existing medical agents remain confined to static architectures or manually curated updates. In contrast, our proposed framework empowers an agent to autonomously identify repeated multi-step sequences in patient care and encapsulate them as new tools. In this self-evolving, experience-driven design, a clinical agent incrementally builds an expanding toolkit of composite workflows drawn directly from real-world clinical patterns~\cite{huang2025biomni,wu2025evolver}.

\begin{figure}[!t]
  \centering
  \includegraphics[width=\textwidth]{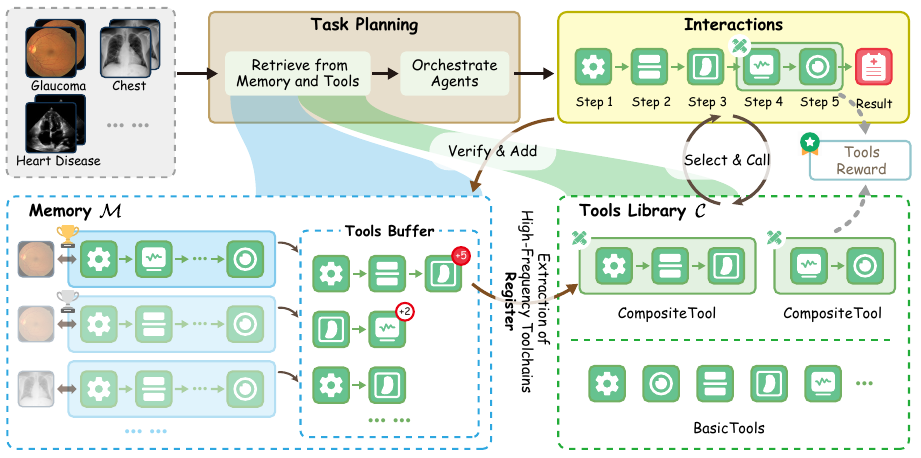}
  \caption{Pipeline of proposed MACRO.  For each input image and multi-step trajectory, MACRO first retrieves relevant experiences from a memory store using image feature similarity. At every step, the model generates a response; if the generated tool sequence contains any registered composite tool $\mathbf{c} \in \mathcal{C}$, a positive reward is assigned. Tool calls are executed, and their results are appended to the evidence, informing subsequent steps. At the end of the trajectory, if the final answer matches the ground truth, the entire tool sequence is stored back into $\mathcal{M}$, and its contiguous subsequences increment the frequency counts in the composite registry $\mathcal{C}$, enabling online discovery of new composite tools.}
  \label{fig:pipeline}
\end{figure}

\section{Methods}
We present a self-evolving agent framework that formulates medical tool-integrated reasoning as sequential decision-making under partial observability (as shown in Fig.~\ref{fig:pipeline}).
The core idea is that the agent not only learns \emph{which} tools to invoke, but also \emph{discovers and internalizes} reusable multi-step procedures from its own successful interactions.
We first formalize the problem (\S~\ref{sec:formulation}), then describe the three pillars of proposed MACRO in this section: experience-grounded memory (\S~\ref{sec:memory}), composite tool discovery (\S~\ref{sec:composite}), and a two-stage policy optimization combining supervised warm-start with reinforcement-based refinement (\S~\ref{sec:optimization}).

\subsection{Problem Formulation}
\label{sec:formulation}

We model the agent--environment interaction as a partially observable Markov decision process (POMDP)
$\mathcal{P}=(\mathcal{S},\mathcal{A},\mathcal{O},T,R,\gamma)$.
Here, the environment state is augmented with an external experience memory,
$\tilde{s}_t=(s_t,\mathcal{M}_t),$
where $s_t$ is the latent clinical/environment state and $\mathcal{M}_t$ is the memory store accumulated from successful episodes.
Accordingly, $T\!: (\mathcal{S}\times\mathfrak{M})\!\times\!\mathcal{A}\!\to\!\Delta(\mathcal{S}\times\mathfrak{M})$ is the transition over both environment and memory states, and $R\!: (\mathcal{S}\times\mathfrak{M})\!\times\!\mathcal{A}\!\to\!\mathbb{R}$ is the reward function.
At each step, the agent only observes a partial view of $\tilde{s}_t$.
Specifically, the agent receives a partial observation as $o_t \;=\; \bigl(\, x,\; h_{<t},\; e_{<t},\; m_t \,\bigr)$,
where $x$ is the input medical image, $h_{<t}$ is the action history, $e_{<t}$ is accumulated tool execution feedback, and
$m_t=\mathrm{retrieve}(\mathcal{M}_t, x, h_{<t}, e_{<t})$
denotes context retrieved from the current memory state (defined in~\S~\ref{sec:memory}).
% \begin{equation}\label{eq:obs}
%   o_t \;=\; \bigl(\, x,\; h_{<t},\; e_{<t},\; m_t \,\bigr)
% \end{equation}

\noindent\textbf{Action space with evolving abstractions.}
The policy $\pi_\theta(a_t\mid o_t)$ selects from an action space $\mathcal{A}_t = \mathcal{T} \cup \mathcal{C}_t \cup \{\texttt{answer}\}$,
where $\mathcal{T}$ is the set of atomic tool calls, $\mathcal{C}_t$ is a dynamically growing set of \emph{composite tools} (multi-step procedures discovered during training; see~\S~\ref{sec:composite}), and \texttt{answer} terminates the episode with a diagnostic output.

\noindent\textbf{Transition and reward.}
When a tool action $a_t$ is selected, the environment executes it and returns the result $r_t$; the context is updated as Eq.~\ref{eq:transition}.
\begin{equation}\label{eq:transition}
  e_{t+1} = e_{<t} \cup \{(a_t, r_t)\}, \qquad h_{t+1} = h_{<t} \cup \{a_t\}
\end{equation}
At episode end, memory evolves as Eq.~\ref{eq:mt1}, 
where $\Delta(\tau_t)$ extracts storable entries from the rollout. We employ a sparse, task-aligned reward during RL: the agent receives a positive signal when its rollout contains validated composite procedures. Final prediction correctness is used as an episode-level quality gate for updating memory and composite registries, encouraging both structured orchestration and task reliability.
\begin{equation}\label{eq:mt1}
\mathcal{M}_{t+1}=\begin{cases}
\mathcal{M}_t \cup \Delta(\tau_t), & \hat{y}=y\\
\mathcal{M}_t, & \text{otherwise}
\end{cases}
\end{equation}

\subsection{Experience-grounded Memory}
\label{sec:memory}

\subsubsection{Off-policy Trajectory Bootstrap.}

To avoid cold-start exploration in the combinatorially large tool space, we first collect high-quality demonstration trajectories from a strong teacher VLM $\mathcal{M}_{\mathrm{teach}}$ (DeepSeek).
For each training sample $x_i$, the teacher performs multi-step reasoning with tool execution as Eq.~\ref{eq:teacher}, 
yielding a trajectory $\tau_i = (o_1, a_1, r_1, \ldots, o_{T_i}, a_{T_i})$.
These trajectories serve dual purposes: they provide supervision for behavior cloning (\S~\ref{sec:sft}) and seed the experience memory below.
\begin{equation}\label{eq:teacher}
  a_t \sim \mathcal{M}_{\mathrm{teach}}\!\bigl(\,\cdot \mid \mathrm{prompt}(x_i, o_t)\bigr), \quad t = 1, \ldots, T_i
\end{equation}

\subsubsection{Memory Store and Retrieval.}

We maintain a memory buffer $\mathcal{M}$ that stores fragments of successful interactions.
Each entry takes the form $m = (\mathbf{p},\; \mathbf{t},\; \mathbf{r},\; \mathbf{f})$, 
where $\mathbf{p}$ is the prompt/history context, $\mathbf{t}$ is the tool invocation sequence, $\mathbf{r}$ is a summarized tool result, and $\mathbf{f}\!\in\!\mathbb{R}^d$ is an image feature vector extracted by a frozen visual encoder.
% \begin{equation}\label{eq:memory_entry}
%   m = (\mathbf{p},\; \mathbf{t},\; \mathbf{r},\; \mathbf{f}),
% \end{equation}

\paragraph{Image-conditioned retrieval.}
Given a query image with feature $\mathbf{f}_q$, we rank memory entries by cosine similarity as in Eq.~\ref{eq:retrieval}, 
and inject the top-$k$ retrieved entries into the prompt as in-context demonstrations.
This provides visually grounded guidance: the agent observes how similar cases were previously solved before committing to a tool plan.
\begin{equation}\label{eq:retrieval}
  \mathrm{sim}(\mathbf{f}_q, \mathbf{f}_m) = \frac{\mathbf{f}_q^\top \mathbf{f}_m}{\|\mathbf{f}_q\|\;\|\mathbf{f}_m\|}
\end{equation}

\paragraph{Online memory update.}
After each training episode, if the final prediction is correct ($\hat{y}_i\!=\!y_i$), we decompose the trajectory into step-level entries and insert them into $\mathcal{M}$:
\begin{equation}\label{eq:mem_update}
  \mathcal{M} \;\leftarrow\; \mathcal{M} \;\cup\; \bigl\{(\mathbf{p}_i^{(j)},\, \mathbf{t}_i^{(j)},\, \mathbf{r}_i^{(j)},\, \mathbf{f}_i) \bigr\}_{j=1}^{K_i}
\end{equation}

\subsection{Composite Tool Discovery}
\label{sec:composite}

While memory retrieval provides soft guidance via in-context examples, composite tool discovery creates \emph{hard} action abstractions that the agent can invoke as single units.

\noindent\textbf{Frequent subsequence mining.}
From the collection of successful trajectories, we extract tool invocation sequences
$\mathbf{s}^{(i)} = (t_1^{(i)}, t_2^{(i)}, \ldots, t_{L_i}^{(i)})$
and mine contiguous subsequences that recur across samples.
A candidate subsequence $\mathbf{c}$ is registered as a composite tool when its frequency exceeds a threshold~$\tau$, $\mathcal{C} = \bigl\{\,\mathbf{c} \;\big|\; \mathrm{freq}(\mathbf{c}) > \tau \,\bigr\}$. 
\noindent where $\mathcal{C}$ is updated throughout training: new composites may be registered as the policy improves and discovers novel successful patterns, making the action space itself a function of accumulated experience.
% \begin{equation}\label{eq:composite}
%   \mathcal{C} = \bigl\{\,\mathbf{c} \;\big|\; \mathrm{freq}(\mathbf{c}) > \tau \,\bigr\}
% \end{equation}

% \begin{algorithm}[t]
% \caption{Compact Two-Stage Training with Memory and Composite Tools}
% \label{alg:two-stage}
% \begin{algorithmic}[1]
% \Require Dataset $\mathcal{D}$, model $\pi_\theta$, memory $\mathcal{M}$, composite registry $\mathcal{C}$.
% \State Initialize $\mathcal{M} \gets \varnothing$, $\mathcal{C} \gets \varnothing$.
% \For{each training sample $(x_i, \tau_i)$}
%   \State Retrieve memory context, build prompt, and train $\pi_\theta$ with SFT objective $\mathcal{L}_{\mathrm{SFT}}$.
%   \State Roll out tool calls, append successful trajectories to $\mathcal{M}$.
%   \State Mine frequent contiguous tool subsequences from successful trajectories and update $\mathcal{C}$.
% \EndFor
% \For{each GRPO epoch}
%   \State Roll out $G$ trajectories per sample with memory retrieval and tool execution.
%   \State Assign reward $R=1$ if any registered composite in $\mathcal{C}$ is invoked, else $R=0$.
%   \State Compute group-normalized advantages and update $\theta$ via Eq.~(\ref{eq:grpo}).
%   \State Add newly successful trajectories to $\mathcal{M}$ and refresh $\mathcal{C}$.
% \EndFor
% \end{algorithmic}
% \end{algorithm}

\subsection{Two-Stage Policy Optimization for Self-Evolving}
\label{sec:optimization}

\subsubsection{Stage~1: Supervised Cold Start.}
\label{sec:sft}

We initialize $\pi_\theta$ on teacher demonstrations $\mathcal{D}=\{\tau_i\}_{i=1}^N$ by minimizing $\mathcal{L}_{\mathrm{SFT}}(\theta)$ in Eq.~\ref{eq:sft}.
where $a_t^*$ is the teacher action and $\hat{o}_t^{(i)}$ is built from the student's (we employ Qwen2.5-VL-3B-Instruct as the base model for training) own execution context: at each step, the student generates $\hat{a}_t\!\sim\!\pi_\theta$ (without gradient) and executes it in the environment, so that the accumulated feedback $\hat{e}_t$ reflects the student's behavioral distribution rather than the teacher's. This mitigates the exposure bias of standard behavior cloning.
After each episode with a correct final prediction ($\hat{y}_i\!=\!y_i$), the resulting trajectory is used to update both the memory store $\mathcal{M}$ and the composite registry $\mathcal{C}$, enabling the agent to discover reusable procedures before reinforcement learning begins.
\begin{equation}\label{eq:sft}
  \mathcal{L}_{\mathrm{SFT}}(\theta)
  = -\sum_{i=1}^{N}\sum_{t=1}^{T_i}
    \log\,\pi_\theta\!\bigl(a_t^{*} \mid \hat{o}_t^{(i)}\bigr)
\end{equation}

\subsubsection{Stage~2: GRPO-based Reinforcement for Composite Utilization.}
\label{sec:grpo}

After a supervised warm-start, the policy can use atomic tools reliably but may under-utilize the discovered composites.
We therefore apply Group Relative Policy Optimization (GRPO)~\cite{shao2024deepseekmath} to reinforce structured tool orchestration.

For each input $x_i$, we sample a group of $G$ on-policy trajectories $\{\tau_i^{(g)}\}_{g=1}^G$ from $\pi_\theta$.
The per-trajectory reward $R_i^{(g)}$ is $1$ if the rollout contains any registered composite $\mathbf{c}\!\in\!\mathcal{C}$, and $0$ otherwise.
We compute group-normalized advantages:
\begin{equation}\label{eq:advantage}
  A_i^{(g)} = \frac{R_i^{(g)} - \mu_i}{\sigma_i + \epsilon}, \qquad
  \mu_i = \frac{1}{G}\sum_{g=1}^{G} R_i^{(g)}, \qquad
  \sigma_i = \sqrt{\frac{1}{G}\sum_{g=1}^{G}\bigl(R_i^{(g)} - \mu_i\bigr)^2}
\end{equation}

The policy is then updated by Eq.~\ref{eq:grpo}, 
where $\rho_t = \pi_\theta(a_t\mid o_t)\,/\,\pi_{\mathrm{ref}}(a_t\mid o_t)$ is the importance ratio and $\pi_{\mathrm{ref}}$ is the reference policy (frozen after Stage~1).
\begin{equation}\label{eq:grpo}\small
  \mathcal{L}_{\mathrm{GRPO}}(\theta)
  = -\mathbb{E}_{i,g,t}\!\Bigl[\,
    \min\!\bigl(\rho_t\, A_i^{(g)},\;
    \mathrm{clip}(\rho_t, 1\!-\!\epsilon, 1\!+\!\epsilon)\, A_i^{(g)}\bigr)
  \,\Bigr]
  + \beta\, D_{\mathrm{KL}}\!\bigl(\pi_\theta \,\|\, \pi_{\mathrm{ref}}\bigr)
\end{equation}

During GRPO rollouts, tool calls are executed to maintain environment-consistent transitions, and trajectories with correct final predictions continue to enrich both $\mathcal{M}$ and $\mathcal{C}$.
\subsubsection{Reward Function.}
We define a sparse reward function \(R(s_t, a_t)\) to incentivize the agent's usage of discovered composite tools during the reinforcement learning stage (Stage~2). Let \(\mathcal{C}\) denote the set of registered composite tools, where each composite tool \(\mathbf{c} \in \mathcal{C}\) is an ordered tuple of atomic tools \(\mathbf{c} = (c_1, c_2, \dots, c_L)\) that has been observed frequently in successful trajectories. At each step \(t\), after the agent generates an action \(a_t\), we parse the action to extract the sequence of tool names invoked within that step (if any). Let \(\mathbf{g}_t\) be this list of tool names. If \(\mathbf{g}_t\) contains any contiguous subsequence that exactly matches a composite tool \(\mathbf{c} \in \mathcal{C}\), i.e., if \(\exists \mathbf{c} \in \mathcal{C}\) such that \(\mathbf{c}\) is a contiguous subsequence of \(\mathbf{g}_t\) (denoted \(\mathbf{c} \sqsubseteq \mathbf{g}_t\)), then a positive reward is assigned as Eq.~\ref{eq:rt}. 
This reward encourages the agent to invoke composite tools as a single conceptual unit, thereby promoting more efficient and structured reasoning.
\begin{equation}
    R_t = \begin{cases}
1, & \text{if } \exists \mathbf{c} \in \mathcal{C} \text{ with } \mathbf{c} \sqsubseteq \mathbf{g}_t\\
0, & \text{otherwise}
\end{cases}
\label{eq:rt}
\end{equation}

In addition to this per-step reward, we also leverage a trajectory-level success signal at the end of each sample: if the model's final answer matches the ground-truth label, we consider the entire tool sequence \(\mathbf{s} = (t^{(1)}, t^{(2)}, \dots, t^{(K)})\) as successful and use it to update the memory store \(\mathcal{M}\) and the composite registry \(\mathcal{C}\). This delayed success signal provides a global credit assignment and enables the discovery of new composite patterns, but it does not directly affect the policy gradient updates in Stage~2. The combination of step-level composite rewards and trajectory-level success feedback allows the agent to progressively refine its multi-step reasoning while automatically acquiring reusable action abstractions.

\noindent %Algorithm~\ref{alg:two-stage} summarizes the full two-stage procedure.
The proposed framework unifies retrieval-augmented experience replay, automated action abstraction, and group-relative policy optimization within a single POMDP-grounded training loop.
The agent's action space, memory, and policy co-evolve: each successful interaction strengthens all three, enabling continual capability growth---a property critical for deployment in diverse and evolving clinical environments.

\section{Experiments}
We compare proposed MACRO with VLMs, task-specific models and medical agentic systems in Section~\ref{E1}. Ablation studies and in-depth analyses are presented in Section~\ref{E2} to demonstrate the effectiveness of our approach. 

\subsection{Experimental Setup}
% 扩充详细数据集情况
% \subsubsection{Datasets.}
We conduct experiments on three datasets with increasingly challenging settings. The
REFUGE2 dataset~\cite{fang2022refuge2} is used for glaucoma diagnosis, and the MITEA dataset~\cite{zhao2023mitea} for heart disease diagnosis (e.g., dilated cardiomyopathy, amyloidosis), RAM-W600~\cite{yang2025ram} dataset is used for bone erosion (BE)  diagnosis using conventional radiography (CR), all of which are suited for evaluating in-depth diagnostic reasoning. 
% \subsubsection{Evaluation Metrics}
% Several metrics are used to evaluate performance: for REFUGE2, MITEA datasets, we report BACC and the F1 score. For the RAM-W600 dataset, we report ACC, SEN, SPC, PRE. The best results are highlighted in bold, and the second-best are underlined.
% \subsubsection{Implementation Details.}

The base model is Qwen2.5-VL-3B-Instruct, a 3B parameter vision-language model. To enable parameter-efficient fine-tuning, we apply LoRA (Low-Rank Adaptation) with rank \(r=8\), LoRA alpha \(\alpha=32\), and dropout rate \(0.1\). The model is optimized using AdamW with a learning rate of \(5\times10^{-5}\) for the supervised stage and \(5\times10^{-6}\) for the GRPO stage. 
% We conduct all experiments on a NVIDIA GeForce RTX 4090 GPU. 

\begin{table}[!t]
\centering
\caption{Comparison with general VLMs and medical agentic systems on REFUGE2~\cite{fang2022refuge2}, MITEA~\cite{zhao2023mitea}. The best results in each column are highlighted in \textbf{bold}, and the second-best values are \underline{underlined}.}
\label{tab:comparison}
\setlength{\tabcolsep}{12pt}
\newcommand{\gray}{\cellcolor{gray!8}}
\resizebox{\linewidth}{!}{
\begin{tabular}{cl cc cc}
\toprule
\multirow{2.5}{*}{\textbf{Type}} & \multirow{2.5}{*}{\textbf{Method}} & \multicolumn{2}{c}{\textbf{REFUGE2}~\cite{fang2022refuge2}} & \multicolumn{2}{c}{\textbf{MITEA}~\cite{zhao2023mitea}} \\ 
\cmidrule(lr){3-4} \cmidrule(lr){5-6} 
 & & BACC $\uparrow$ & F1 $\uparrow$ & BACC $\uparrow$ & F1 $\uparrow$ \\ 
\midrule
\multirow{6}{*}{General}
& \gray GPT-4o~\cite{achiam2023gpt}          & \gray 56.4 & \gray 21.1 & \gray 56.8 & \gray 28.1 \\
& Janus-Pro-7B~\cite{chen2025janus}    & 53.4 & 13.3 & 52.3 & 10.7 \\
& \gray LLaVA-Med~\cite{li2023llava}        & \gray 50.0 & \gray ~0.0 & \gray 50.0 & \gray ~0.0 \\
& BioMedClip~\cite{zhang2023biomedclip}      & 58.1 & 21.3 & 47.0 & 37.8 \\
& \gray Qwen2.5-7B-VL~\cite{wang2024qwen2}   & \gray 54.3 & \gray 16.3 & \gray 50.0 & \gray ~0.0 \\
& InternVL2.5-8B~\cite{chen2024internvl}  & 51.8 & 13.8 & 49.7 & ~3.6 \\ 
\midrule
\multirow{5}{*}{Agentic}
& \gray MedAgents~\cite{tang2024medagents}       & \gray 52.1 & \gray ~8.9 & \gray 51.1 & \gray 15.9 \\
& MMedAgent~\cite{li2024mmedagent}       & 52.4 & 16.3 & 55.0 & 26.7 \\
& \gray MDAgents~\cite{kim2024mdagents}         & \gray 56.8 & \gray 22.2 & \gray 57.2 & \gray 30.3 \\
& MedAgent-Pro~\cite{wang2025medagent}    & \underline{90.4} & \underline{76.4} & \textbf{77.8} & \underline{72.3} \\
& \cellcolor{blue!10}{MACRO~(Ours)} & \cellcolor{blue!10}\textbf{92.7} & \cellcolor{blue!10}\textbf{80.3} & \cellcolor{blue!10}\underline{77.2} & \cellcolor{blue!10}\textbf{74.9} \\
\bottomrule
\end{tabular}
}
% \enlargethispage{12pt}
\end{table}

\subsection{Comparison Experiments}
\label{E1}
\subsubsection{Comparison with General VLMs.}
We compare proposed MACRO against a comprehensive set of state-of-the-art vision-language models (VLMs) and medical agentic systems under identical experimental settings, including BioMedClip~\cite{zhang2023biomedclip}, GPT-4o~\cite{achiam2023gpt}, LLaVA-Med~\cite{li2023llava}, Janus~\cite{chen2025janus}, Qwen~\cite{wang2024qwen2}, InternVL~\cite{chen2024internvl}. To ensure fair evaluation, for the MITEA dataset (3D echocardiography), we follow the same protocol as MedAgent-Pro~\cite{wang2025medagent} by randomly selecting three slices from each volume as visual input. As reported in Table~\ref{tab:comparison}, our proposed MACRO   framework outperforms all existing VLMs across both glaucoma and heart disease diagnosis tasks. Specifically, compared to Qwen, the proposed MACRO achieves absolute improvements of 38.4\% in balanced accuracy (BACC) and 64.0\% in F1 score for glaucoma, and 27.2\% in BACC and 74.9\% in F1 score for heart disease. These results highlight the effectiveness of the proposed MACRO workflow in tackling complex medical diagnosis tasks. By integrating the self-evolving composition of vision tools into the inference process, proposed MACRO enables precise metric computation and enhanced diagnostic performance, effectively addressing the limitations of existing vision-language models.

% 前两数据集结果
% \begin{table}[!t]
% \centering
% \caption{Comparison with general VLMs and medical agentic systems on REFUGE2, MITEA and RAM-W600.}
% \label{tab:comparison}
% \resizebox{\linewidth}{!}{
% \begin{tabular}{l c c c c}
% \toprule
% \multirow{2}{*}{Method} & \multicolumn{2}{c}{Glaucoma} & \multicolumn{2}{c}{Heart Disease} \\ \cmidrule(lr){2-3} \cmidrule(lr){4-5} 
%  & BACC & F1 & BACC & F1 \\ \midrule
% GPT-4o [12] & 56.4 & 21.1 & 56.8 & 28.1   \\
% Janus-Pro-7B [13] & 53.4 & 13.3 & 52.3 & 10.7  \\
% LLaVA-Med [4] & 50.0 & 0.0 & 50.0 & 0.0  \\
% BioMedClip [24] & 58.1 & 21.3 & 47.0 & 37.8 \\
% Qwen2.5-7B-VL [81] & 54.3 & 16.3 & 50.0 & 0.0\\
% InternVL2.5-8B [82] & 51.8 & 13.8 & 49.7 & 3.6\\ \midrule
% MedAgents [30] (ACL'24) & 52.1 & 8.9 & 51.1 & 15.9\\
% MMedAgent [32] (EMNLP'24) & 52.4 & 16.3 & 55.0 & 26.7\\
% MDAgents [29] (NeurIPS'24) & 56.8 & 22.2 & 57.2 & 30.3\\
% MedAgent-Pro (ICLR'26) & 90.4 & 76.4 & 77.8 & 72.3\\
% Ours &92.7  &80.3  &77.2  &74.9  \\
% \bottomrule
% \end{tabular}}
% \end{table}

\subsubsection{Comparison with Medical Agentic Systems}
% 前两数据集结果
We further compare proposed MACRO against state-of-the-art medical agentic frameworks, including MedAgents~\cite{tang2024medagents}, MMedAgent~\cite{li2024mmedagent}, MDAgents~\cite{kim2024mdagents}, and MedAgent-Pro~\cite{wang2025medagent}. As shown in Table~\ref{tab:comparison}, the proposed MACRO consistently outperforms all these methods across both disease domains. This performance advantage stems from the fact that prior approaches—such as MedAgents, MMedAgent, and MDAgents, are primarily designed for basic question answering or modular tool utilization, lacking the capacity to handle complex multimodal clinical scenarios. While MedAgent-Pro also integrates multiple tools as a strong agentic baseline, our approach further advances the state of the art: we achieve gains of 2.3\% in BACC and 3.9\% in F1 score on glaucoma, and while BACC shows a marginal decrease of 0.6\% on heart disease, F1 score improves by a notable 2.6\%, demonstrating overall superior diagnostic balance. By enabling self-evolving tool composition, the proposed MACRO is better aligned with the inherent domain variability and diagnostic complexity of clinical medicine, thereby delivering more effective and comprehensive decision support in real-world applications.

\begin{table*}[!t]
\centering
% \scriptsize
\caption{BE classification results obtained on the Test set of RAM-W600~\cite{yang2025ram}. The best results in each column are highlighted in \textbf{bold}, and the second-best values are \underline{underlined}.}
\label{tab:be_results}
\newcommand{\gray}{\cellcolor{gray!8}}
\newcommand{\best}{\cellcolor{blue!10}}
\setlength{\tabcolsep}{9pt}
\resizebox{\textwidth}{!}{
\begin{tabular}{lcccccc}
\toprule
\textbf{Method} & BACC$\uparrow$ & F1$\uparrow$ & ACC$\uparrow$ & SEN$\uparrow$ & SPC$\uparrow$ & PRE$\uparrow$ \\
\midrule
% \multirow{5}{*}{\rotatebox{90}{\small General}}
\gray MobileViT~\cite{mehta2021mobilevit}       & \gray \underline{52.64} & \gray 11.85 & \gray 81.42 & \gray 21.06 & \gray 84.23 & \gray 9.31 \\
ResNet~\cite{he2016deep}                  & 51.75 & 10.89 & 78.27 & 23.10 & 80.40 & 7.79 \\
\gray MobileNet~\cite{howard2017mobilenets}         & \gray 47.84 & \gray 10.79 & \gray 74.08 & \gray 17.02 & \gray 78.66 & \gray 9.07 \\
LeViT~\cite{graham2021levit}                   & 49.29 & 6.73  & 84.17 & 8.49  & 90.09 & 8.99 \\
\gray EfficientFormer~\cite{li2022efficientformer}   & \gray 50.63 & \gray \underline{12.40} & \gray 72.04 & \gray \underline{27.90} & \gray 73.37 & \gray 8.82 \\
% \midrule
% \multirow{3}{*}{\rotatebox{90}{\small Med.}}
MedMamba~\cite{yue2024medmamba}                & 50.83 & 6.91  & 86.56 & 8.94  & 92.73 & \underline{11.56} \\
\gray ConvKAN~\cite{bodner2024convolutional}           & \gray 49.26 & \gray 3.49  & \gray \underline{87.42} & \gray 3.82  & \gray \underline{94.70} & \gray 6.56 \\
\best{MACRO~(Ours)}          & \best\textbf{61.75} & \best\textbf{30.00} & \best\textbf{90.01} & \best\textbf{28.50} & \best\textbf{95.00} & \best\textbf{31.61} \\
\bottomrule
\end{tabular}
}
\end{table*}

\subsubsection{Comparison with Task-specific Models.}
Additionally, we compare proposed MACRO against task-specific methods that are explicitly trained on the RAM-W600 dataset. The purpose of this experiment is to test whether MACRO can still remain competitive when facing specialized models optimized for the same task and data distribution. All selected models, including MobileViT~\cite{mehta2021mobilevit}, ResNet~\cite{he2016deep}, MobileNet~\cite{howard2017mobilenets}, LeViT~\cite{graham2021levit}, EfficientFormer~\cite{li2022efficientformer}, MedMamba~\cite{yue2024medmamba}, and ConvKAN~\cite{bodner2024convolutional}, were trained and tested on this specific dataset to ensure a fair comparison of their specialized diagnostic capabilities. Following standard practice, classification performance was quantified using BACC, F1 score, accuracy (ACC), sensitivity (SEN), specificity (SPC), and precision (PRE).

The results in Table~\ref{tab:be_results} show that mainstream models trained on RAM-W600 achieve modest BACC and F1 score (max 52.64\% and 12.40\%), with some (e.g., ConvKAN) obtaining high specificity (94.70\%) but extremely low sensitivity (3.82\%), indicating a strong bias toward the majority class. This reflects the inherent difficulty of BE detection due to class imbalance and subtle radiographic features. In contrast, proposed MACRO   leverages a multimodal agentic framework that dynamically invokes tools (e.g., edge detection, segmentation) tailored to BE characteristics, achieving substantial gains across all metrics. The balanced improvement in sensitivity and precision demonstrates that task-adaptive tool integration effectively captures subtle erosive changes missed by conventional classifiers, mitigating class imbalance and enhancing diagnostic reliability. 

% \begin{table*}[p]
% \centering
% \caption{BE $\&$ nonBE classification results obtained on the Test set.}
% \label{tab: be_results}
% \begin{threeparttable}
% \setlength{\tabcolsep}{2pt} % 临时调整列间距
% \resizebox{0.92\textwidth}{!}{
% \centering
% \begin{tabular}{lcccccc}
% \toprule
% \textbf{Model} & {\makecell{BACC}} & {\makecell{F1-Score}} & {\makecell{ACC}} & {\makecell{SEN}} & {\makecell{SPC}} & {\makecell{PRE}} \\
% \midrule
% MobileViT~\cite{mehta2021mobilevit}       & \underline{52.64} & 11.85 & 81.42 & 21.06 & 84.23 & 9.31 \\
% ResNet~\cite{he2016deep}                  & 51.75 & 10.89 & 78.27 & 23.10 & 80.40 & 7.79 \\
% MobileNet~\cite{howard2017mobilenets}     & 47.84 & 10.79 & 74.08 & 17.02 & 78.66 & 9.07 \\
% LeViT~\cite{graham2021levit}              & 49.29 & 6.73 & 84.17 & 8.49 & 90.09 & 8.99 \\
% EfficientFormer~\cite{li2022efficientformer} & 50.63 & \underline{12.40} & 72.04 & \underline{27.90} & 73.37 & 8.82 \\
% MedMamba~\cite{yue2024medmamba}           & 50.83 & 6.91 & 86.56 & 8.94 & 92.73 & \underline{11.56} \\
% ConvKAN~\cite{bodner2024convolutional}    & 49.26 & 3.49 & \underline{87.42} & 3.82 & \underline{94.70} & 6.56 \\
% Ours                                       & \textbf{61.75} & \textbf{30.00} & \textbf{90.01} & \textbf{28.50} & \textbf{95.00} & \textbf{31.61 }\\
% \bottomrule
% \end{tabular}
% }
% \end{threeparttable}
% \end{table*}
% \clearpage
\subsection{Ablation Study and Detailed Analysis}
\label{E2}
\subsubsection{Effectiveness of the Proposed Key Components.}
We design an ablation study comprising three conditions to evaluate the contribution of each core component: (A) baseline without memory, composite tool discovery, or GRPO; (B) adding only memory retrieval; (C) further enabling composite tool discovery (without GRPO reinforcement); and (D) the full system with memory, composite tool discovery, and GRPO. All other hyperparameters (learning rate, LoRA, training epochs etc.) are kept identical across conditions.
\begin{table}[!t]
\centering
\caption{Ablation on key components, including memory store, composite tool discovery, and GRPO. The best results in each column are highlighted in \textbf{bold}, and the second-best values are \underline{underlined}.}
\label{tab:ablation}
\newcommand{\cmark}{\ding{51}}
\newcommand{\xmark}{\ding{55}}
\newcommand{\gray}{\cellcolor{gray!8}}
\newcommand{\best}{\cellcolor{blue!10}}
\setlength{\tabcolsep}{13pt}
\resizebox{\linewidth}{!}{
\begin{tabular}{ccc cc cc}
\toprule
\multicolumn{3}{c}{\textbf{Components}} & \multicolumn{2}{c}{\textbf{REFUGE2}~\cite{fang2022refuge2}} & \multicolumn{2}{c}{\textbf{MITEA}~\cite{zhao2023mitea}} \\
\cmidrule(lr){1-3} \cmidrule(lr){4-5} \cmidrule(lr){6-7}
Memory & Composite & GRPO & BACC & F1 & BACC & F1 \\
\midrule
\gray \xmark & \gray \xmark & \gray \xmark & \gray 82.8 & \gray 69.7 & \gray 69.2 & \gray 64.8 \\
\cmark & \xmark & \xmark & 83.9 & 75.8 & 72.4 & 65.3 \\
\gray \cmark & \gray \cmark & \gray \xmark & \gray \underline{90.6} & \gray \underline{80.3} & \gray \underline{75.1} & \gray \underline{71.8} \\
\best \cmark & \best \cmark & \best \cmark & \best \textbf{92.7} & \best \textbf{80.3} & \best \textbf{77.2} & \best \textbf{74.9} \\
\bottomrule
\end{tabular}
}
\end{table}

From the Table \ref{tab:ablation}, the ablation study demonstrates that each proposed component contributes positively to the overall system. The baseline model achieves moderate performance (REFUGE2 BACC 82.8~\%, F1 69.7~\%). Adding memory store improves recall and F1, particularly on REFUGE2 (F1 +6.1~\%). Incorporating composite tool discovery without reinforcement yields a substantial leap (REFUGE2 BACC 90.6~\%, F1 80.3~\%), confirming that learning multi-step tool patterns significantly enhances reasoning. Finally, GRPO reinforcement further boosts performance to the highest levels (REFUGE2 BACC 92.7~\%, MITEA BACC 77.2~\%), validating that policy-gradient optimization effectively encourages the adoption of discovered composites. The combination of all three components achieves the best results across both datasets, underscoring their synergistic effect.

% \begin{table}[!t]
% \centering
% \caption{Ablation on key components, including MemoryStore, tool Composite and GRPO.}
% \label{tab:ablation}
% \begin{tabular}{c c c c c c c}
% \toprule
% \multicolumn{3}{c}{Settings} &\multicolumn{2}{c}{Glaucoma} & \multicolumn{2}{c}{Heart Disease} \\
% \cmidrule(lr){1-3}\cmidrule(lr){4-5} \cmidrule(lr){6-7}
% Memory & Composite & GRPO & BACC & F1 & BACC & F1 \\
% \midrule
%  & & & & & & \\
% \checkmark & & & & & & \\
% \checkmark & \checkmark & & & & & \\
% \checkmark & \checkmark & \checkmark &92.7  &80.3 &77.2  &74.9 \\
% \bottomrule
% \end{tabular}
% \end{table}

\begin{figure}[!t]
  \centering
  \includegraphics[width=\textwidth]{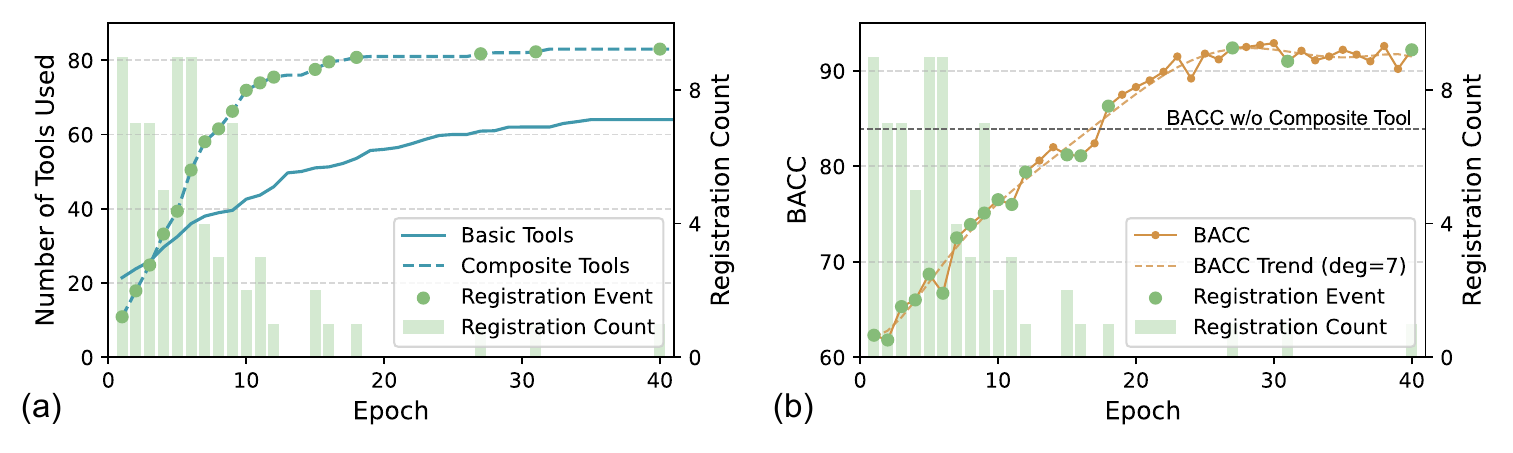}
  \caption{Closed-loop learning in MACRO yields two measurable benefits:
(a) tool complexity decreases as higher-level tools progressively replace multi-step patterns;
(b) As MACRO acquires stronger tools and better demonstrations, performance with abstract tools improves by 8.8\% over the basic tool lib.}
  \label{fig:composite}
\end{figure}

\subsubsection{Analysis of the Number of Registered Composite Tools During Training.}
From Fig.~\ref{fig:composite}~(a), the number of registered composite tools follows a two-phase pattern: rapid growth followed by stabilization. Early in training, effective multi-step patterns are quickly discovered and registered, leading to a sharp increase. GRPO incentivizes usage, boosting frequency. Later, registration saturates as common combinations are covered, and usage stabilizes via reward and memory reinforcement, marking the transition from exploration to exploitation.
From Fig.~\ref{fig:composite}~(b), the growth in registered composites strongly correlates with BACC improvement. Each new composite expands the model's behavioral repertoire, and GRPO reinforcement enables efficient task completion, driving BACC gains.

\subsubsection{Quantitative Analysis}
\begin{figure}[!t]
  \centering
  \includegraphics[width=\textwidth]{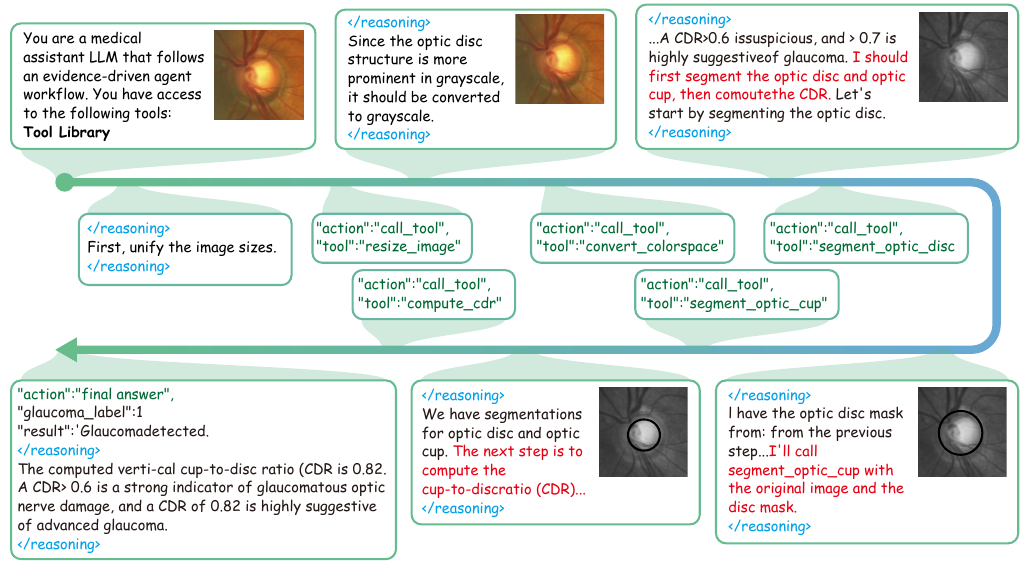}
  \caption{A case study on glaucoma diagnosis, illustrating the detailed workflow within the MACRO framework. The green text represents the call action operations, while the surrounding reasoning reflects the model's explanation and thought process behind each action.}
  \label{fig:workflow}
\end{figure}
In the quantitative analysis phase, specialized tool agents are introduced to perform targeted assessments, thereby bridging the gap between artificial intelligence and clinical practice. The tool set $\mathcal{C}$ comprises various vision models, such as \textit{segment\_optic\_disc}, \textit{segment\_optic\_cup}, and basic image processing tools like \textit{canny\_edge} and \textit{convert\_color\_space}. As illustrated in Fig.~\ref{fig:workflow}, the image is first converted to grayscale, after which specialized segmentation tools are employed to extract the optic cup and optic disc masks. Subsequently, a tool is invoked to compute the cup-to-disc ratio based on the segmentation results—a key indicator for glaucoma diagnosis. This standard workflow for glaucoma diagnosis is also registered as a composite tool within the tool set $\mathcal{C}$.

\section{Discussion and Conclusion}

\textbf{From static orchestration to continual capability growth.}
Our central insight is that the limiting factor of medical agents is no longer raw tool availability, but the ability to \emph{accumulate procedural competence} after deployment. Static tool sets can perform well in controlled settings, yet they fail to keep pace with evolving clinical protocols, data distributions, and task requirements. By distilling recurrent successful trajectories into reusable composite tools, our framework transforms interaction history into an explicit, expandable skill library—shifting the agent from rigid orchestration to continual capability growth, in a manner closer to how clinicians refine their expertise through practice.

% \paragraph{\textbf{Composite tools improve reliability where it matters most.}}
% A notable finding is \emph{where} the gains concentrate. Across datasets, the largest improvements appear in difficult and minority-case settings, where subtle or atypical findings are frequently overlooked by rigid pipelines. This suggests that composite tools do more than compress multi-step prompts: they impose structured verification pathways (e.g., detection $\rightarrow$ localization $\rightarrow$ quantification $\rightarrow$ consistency check) that improve decision reliability under ambiguity and class imbalance. The simultaneous gains in both sensitivity and precision further indicate that the proposed MACRO enhances intrinsic diagnostic utility rather than merely shifting an operating threshold.

\paragraph{\textbf{A practical path toward maintainable clinical deployment.}}
From a translational perspective, this design offers a viable route for continual improvement without repeated end-to-end re-engineering. Newly validated workflows can be absorbed as auditable, versioned primitives, potentially reducing maintenance costs across institutions and enabling safer, more traceable capability updates. This property is especially relevant for real-world deployments where clinical protocols and patient case-mix drift over time.

\paragraph{\textbf{Limitations and future directions.}}
Several limitations warrant discussion. First, composite quality is fundamentally bounded by trajectory verification fidelity; noisy or partially correct traces risk introducing suboptimal skills into the library. Second, transfer to unseen imaging modalities with substantially different visual semantics may still expose failure modes not captured by current benchmarks. Third, our evaluation focuses on predictive metrics; future work should incorporate prospective human-in-the-loop studies assessing calibration, workflow efficiency, and downstream clinical decision impact. Addressing these challenges is essential for translating benchmark gains into trustworthy, real-world adoption.

% \section*{Acknowledgments}

\bibliographystyle{splncs04}
\bibliography{main}
\end{document}